\documentclass[letterpaper, 10 pt, conference]
{ieeeconf}  

\IEEEoverridecommandlockouts                              

\usepackage{color,soul}
\usepackage{hyperref}

\newcommand{\acro}{SynSculptor }
\title{\LARGE \bf
Humanoid Motion Scripting with Postural Synergies
}
\usepackage{amsmath} 
\usepackage{amssymb}
\usepackage{xcolor}
\usepackage{graphicx}

\author{Rhea Malhotra$^{1}$, William Chong, Catie Cuan, and Oussama Khatib%
\thanks{$^{1}$Computer Science,
        Stanford University, Stanford CA USA
        {\tt\small rheamal@stanford.edu}}%
}
\begin{document}

\maketitle
\thispagestyle{empty}
\pagestyle{empty}

\begin{abstract}
Generating sequences of human-like motions for humanoid robots presents challenges in collecting and analyzing reference human motions, synthesizing new motions based on these reference motions, and mapping the generated motion onto humanoid robots. To address these issues, we introduce SynSculptor, a humanoid motion analysis and editing framework that leverages postural synergies for training-free human-like motion scripting. To analyze human motion, we collect 3+ hours of motion capture data across 20 individuals where a real-time operational space controller mimics human motion on a simulated humanoid robot. The major postural synergies are extracted using principal component analysis (PCA) for velocity trajectories segmented by changes in robot momentum, constructing a style-conditioned synergy library for free-space motion generation. To evaluate generated motions using the synergy library, the foot-sliding ratio and proposed metrics for motion smoothness involving total momentum and kinetic energy deviations are computed for each generated motion, and compared with reference motions. Finally, we leverage the synergies with a motion-language transformer, where the humanoid, during execution of motion tasks with its end-effectors, adapts its posture based on the chosen synergy. 
Supplementary material, code, and videos are available at \href{https://rhea-mal.github.io/humanoidsynergies.io/}{https://rhea-mal.github.io/humanoidsynergies.io/}.
\end{abstract}

\section{INTRODUCTION}
The human body is a versatile system with 360 joints driven by nearly 650 muscles. Despite the complexity of everyday control tasks, humans are able to perform such versatile dynamic motions effortlessly. To transfer this intuition onto a humanoid robotic system, adaptable low-dimensional encodings of human-learned strategies are critical for generalized humanoid motion generation. 

Humanoid robots offer the promise of expressive, human-like motion. Motion generation, however, remains notoriously difficult, as the large number of controllable parts in high-DOF robots makes determining optimal movements solely through computationally intense. Real-time motion mapping enables humanoids to directly replicate human movement captured, for control that mirrors human kinematics and style.
To enable deployment in real-world applications, humanoid control representations must generalize across diverse human postures, body types, and stylistic variations. Our approach embeds motion trajectories within subspaces defined by human-derived synergies, providing a scalable framework for generalization and seamless transfer across individuals and tasks. 

We hypothesize that human movement exhibits structured variability, in which a low-dimensional basis of postural synergies orchestrates the coordination of global body motion. Stylistic adjustments then introduce task-specific refinements atop these fundamental synergies. We take advantage of these postural synergies in an intuitive motion-editing interface for dynamic tasks, such as dance, where coordination, expressivity, and adaptability are essential. Our framework integrates with generative models to project motions into style-specific synergy subspaces, enabling real-time personalization without retraining.

Manually manipulating joint angles or hyperparameter tuning for fine adjustments requires complex data collection, often yields unintuitive motions, and can be opaque to end users. Rather, we present the \acro Motion Editor, a lightweight, training-free interface that exposes a compact set of principal‐synergy sliders for real-time humanoid choreography. Users can blend and sequence prototypical moves, adjust step size, duration, and interpolation style, and immediately observe the resulting full-body pose and dynamics. This transforms low-dimensional synergy control into an intuitive, developer-friendly tool for expressive, free-space motion programming.

\begin{figure*}[h]
    \centering
    \includegraphics[width=0.99\linewidth]{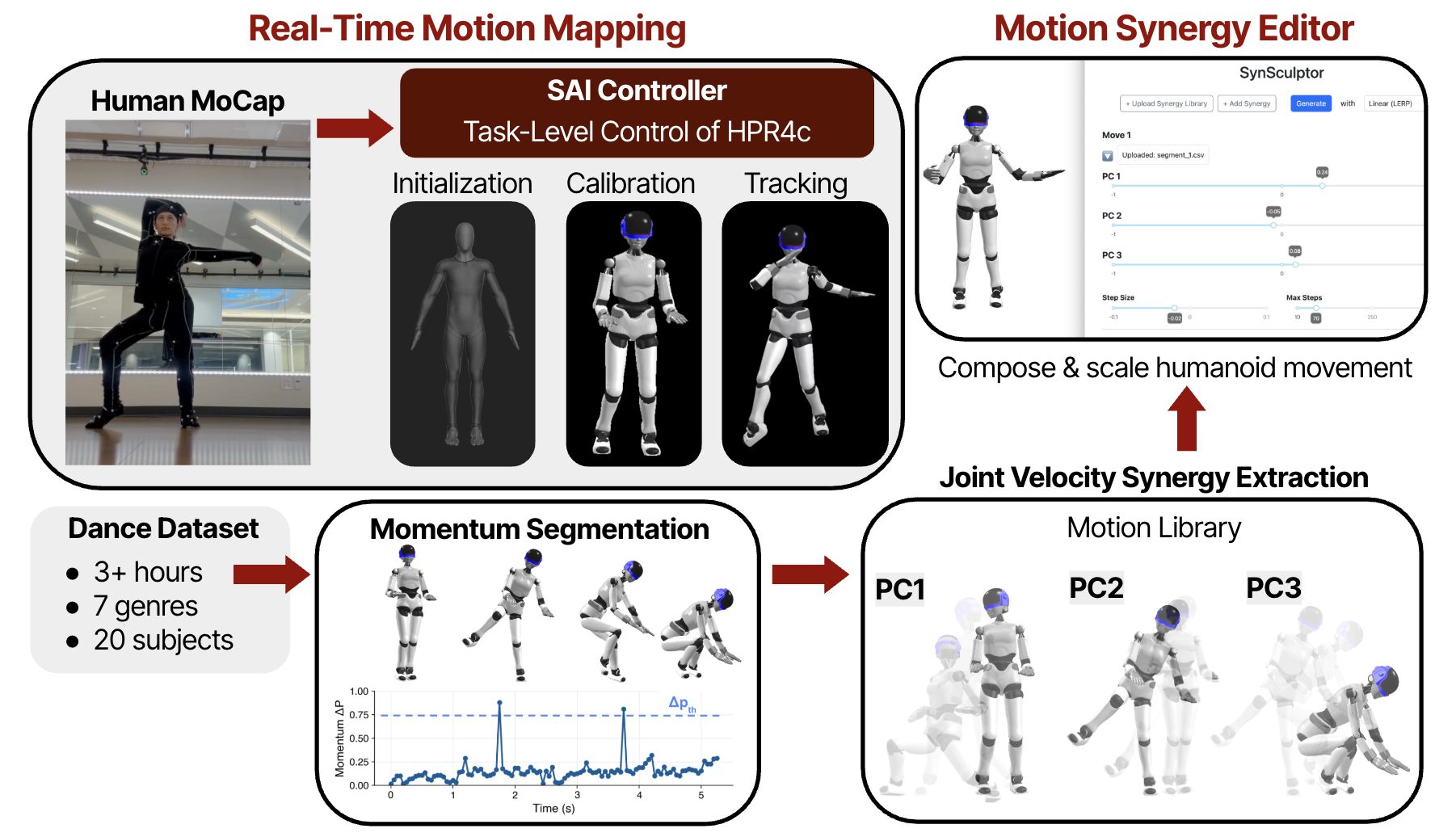}
    \caption{\textbf{Humanoid Motion Mapping-Editing End-to-end Pipeline:} We segment over three hours of MoCap data into distinct “moves” by thresholding whole‐body momentum to extract a compact synergy basis. SynSculptor allows tuning synergies to compose new humanoid trajectories, along with our real-time motion mapping OpenSai controller on a HPR4c humanoid robot.}
    \label{fig:fig1}
\end{figure*}

To this end, our contributions are as follows:

\begin{itemize}
  \item Real-time mapping of OptiTrack motion-capture data onto a simulated, floating-base humanoid with operational space control.
  \item Momentum-based segmentation and PCA-driven extraction of joint‐velocity synergies, enabling compact reconstruction and real-time modulation of movements.
  \item Construction of synergy libraries from human motion data for integration with the SynSculptor Motion Editor 
  \item On-the-fly fine-tuning of MotionGPT outputs via null-space synergy projection, delivering personalized, training-free humanoid postural adaptation.
\end{itemize}

\section{RELATED WORK}
\subsection{Human-to-Humanoid Motion Mapping}
Humanoid robots possess numerous degrees of freedom (DoFs), posing a challenge for mapping human motions onto them. Early work on human-to-humanoid motion transfer explored how methods for mapping upper body motions and balance and angular momentum adjustment strategies restrict joint ranges and velocities, achieving qualitative imitation but limited dynamic realism \cite{pollard2002adapting, natsuk2005whole}. Real-time whole-body imitation via inverse kinematics enabled free-space behaviors like dance and acrobatics \cite{montecillo2010realtime}, and hierarchical control frameworks reproduced complex human-demonstrated motions \cite{arduengo2021,riley2000}. However, these direct kinematic mappings struggle to scale to stylistically diverse or more complex free-space motion repertoires.

Learning‐based methods have enabled dynamic whole‐body behaviors (e.g., walking, jumping, boxing) but demand vast amounts of data and generalize poorly to unseen tasks \cite{shi2020, hester2010, he2024}. High‐dimensional joint action spaces and sample inefficiency remain key barriers to real‐time, robust deployment on full‐size humanoids \cite{harnooja2024}.

\subsection{Stylistic Motion Realism and Expressive Behavior}
Beyond basic motion replication, generating stylistically rich and expressive behaviors is crucial for humanoid robots intended to operate in human environments. Recent work has incorporated style transfer techniques, emotion-conditioned motion, and expressive policy learning \cite{ma2025, hopkins2024}.
Neural Policy Style Transfer transfers emotional motion styles to robots using autoencoder-based feature extraction and reinforcement learning \cite{fernandez2023}. Others prioritize upper-body imitation while relaxing constraints on the legs to enable diverse behaviors such as dancing and shaking hands \cite{cheng2024}. 

Prior methods emphasize task‐specific decomposition or are confined to narrow domains. Our \acro framework delivers diverse, full‐body, real‐time, style‐controllable humanoid motion with no per–task retraining required.

\subsection{Structured Variability in Human Motor Control}
Human motor control inherently leverages structured variability, coordinating complex motions through low-dimensional synergies across joints and muscles. Neuroscientific studies have demonstrated the stability and generality of synergy structures across repeated motor tasks and individuals \cite{katyara2021, merel2018, nenchev2022}. Extending these insights to humanoid motion offers the potential to simplify motion generation and support stylistic adaptability through compact, task-relevant representations.

Postural grasp synergies were first identified in joint angles of the human hand, exhibiting strong correlations during static grasps for compact representation in a low-dimensional synergy space \cite{santello1998, leo2016, thomik2015, brown2007, catalano2012}. This concept has since influenced the design and control of many robotic hands \cite{rosmarin2008, bicchi2011}, facilitating intuitive and efficient representations of complex grasps through synergistic eigenmotions and structured postural spaces \cite{romero2013, konnaris2016}. While these approaches have proven effective in hand-centric tasks, their application to full-body humanoid control remains largely underexplored due to the dynamic complexity and higher dimensionality involved.

Kinematic synergies coupling individual joints into low-dimensional control patterns have been leveraged for humanoid gait control \cite{moiseev2022, hauser2007}. Structured variability emerges in whole-body pose dynamics, with stable, identifiable patterns persisting under covariate conditions \cite{gupta2021, boe2021}. Complex coordinated motions further maintain consistent temporal synergy structures, differing primarily by phase shifts across realizations \cite{brambilla2023}. These findings motivate the use of low-dimensional synergy-based models for humanoid control, reducing motion generation and adaptation complexity.

 \subsection{Robotic Motion Editors}
 Designing expressive and physically feasible robot motions remains challenging due to the high dimensionality and physical constraints of humanoid systems. Various motion-editing frameworks have sought to simplify this task through graphical user interfaces, playback-based editing, and procedural animation \cite{shin2006, nakaoka2004}. However, existing tools often depend on manual keyframing, rigid interpolation methods, or hardware-specific assumptions, limiting their scalability to complex, dynamic, or stylistically diverse behaviors \cite{kuroki2003motion}.

Manual keyframing, where joint positions are specified at discrete frames and interpolated between, fails to capture high-frequency or task-critical dynamics and often introduces artifacts in complex motions \cite{tsai2014, pierris2009, lu2024}. 

Existing humanoid motion generation methods still require (i) extensive task- or environment-specific tuning and (ii) struggle to guarantee both physical feasibility and stylistic variety. This motivates our low-dimensional, synergy-based framework, which provides a style-controllable motion representation without per-task retraining. \acro employs a constraint‐consistent, task‐oriented whole‐body control framework for the simulated HPR4c humanoid \cite{khatib2022} for dynamically-consistent motion execution.

\section{APPROACH}
 Our approach seeks to extract whole-body human synergies from reference human motions to generate human-like movements for humanoid robots. First, we develop a real-time motion mapping framework to map human motion onto a dynamically-simulated, floating-base humanoid robot using the constraint-consistent operational space control framework. Second, we introduce a momentum-based segmentation method for extracting trajectory segments from motion mapping to extract humanoid joint velocity synergies. These extracted synergy libraries form the foundation of SynSculptor, our motion editing tool designed for humanoid motion scripting and animation. While our framework does not enforce physical contact constraints, it is well-suited for generating plausible, expressive humanoid motions as a basis for human-like motion generation. Lastly, we demonstrate integration of language-conditioned outputs from MotionGPT for humanoid animation with stylistic synergies.

\subsection{Mapping Human Motions onto Humanoid Robots}
We present a framework for real-time humanoid motion mapping, leveraging Motion Capture (MoCap) data from OptiTrack (NaturalPoint, Inc., USA) to control a simulated floating-base humanoid (HPR4c). The approach integrates motion tracking, kinematic re-targetting, and energetic evaluation to analyze humanoid motion (Figure \ref{fig:fig1}). OptiTrack tracks markers corresponding to key anatomical landmarks, including the head, hands, elbows, upper torso, pelvis, and feet. The motion data is then streamed to a real-time robot controller to execute torque commands to a simulated humanoid robot using
\href{github.com/manips-sai-org/OpenSai}{OpenSai} at 1 kHz. 

We use a 41-marker OptiTrack baseline skeleton to track the head, hands, elbows, torso, pelvis, and feet via the NatNetLinux interface. The user first adopts a reference stance to establish initial reference frames for each landmark. At 100 Hz, we stream each landmark’s pose and velocity (relative to its initial frame) as controller inputs.

We compute commanded joint torques using the constraint-consistent operational space control framework \cite{khatib2003}. Table \ref{tab:hierarchy} shows the stack-of-tasks used within the framework, where each control task is either the 6 DoF pose task or the 3 DoF orientation task. To control the remaining DoFs after the pose and orientation tasks, a joint posture task is used as the lowest priority task. 

\begin{table}[htbp]
\caption{Stack-Of-Tasks Hierarchy}
\centering
\begin{tabular}{|c|c|c|}
\hline
\textbf{Priority Level} & \textbf{Anatomical Landmark} & \textbf{Task} \\
\hline
1 & Pelvis & Pose \\
1 & Right Foot & Pose \\
1 & Left Foot & Pose \\
1 & Right Hand & Pose \\
1 & Left Hand & Pose \\
1 & Head & Orientation \\
2 & Upper Torso & Orientation \\
2 & Right Elbow & Orientation \\
2 & Left Elbow & Orientation \\
3 & Posture & Joint \\
\hline
\end{tabular}
\label{tab:hierarchy}
\end{table}

Given the stack-of-tasks hierarchy, the corresponding whole-body task representation and task Jacobian are represented with $\vartheta_{\otimes}$ and $J_{\otimes}$, where $\vartheta_{\otimes}$ denotes the instantaneous velocity of the whole-body task. Given the equations of motion of a robot:
\begin{align}
    A(\mathbf{q}) \ddot{\mathbf{q}} + b(\mathbf{q}, \dot{\mathbf{q}}) + g(\mathbf{q}) = \Gamma 
\end{align}
where $A$ is the joint space mass matrix, $b(\mathbf{q}, \dot{\mathbf{q}})$ are the Coriolis and Centrifugal forces, $g(\mathbf{q})$ is the joint space gravity vector, and $\Gamma$ is the vector of joint torques, then the whole-body task space dynamics is obtained by projecting the equations of motion using the dynamically-consistent inverse task Jacobian $\overline{J}_{\otimes}$:
\begin{align}
    \Lambda_{\otimes} \dot{\vartheta}_{\otimes} + \mu_{\otimes} + p_{\otimes} = F_{\otimes} 
\end{align}
where
\begin{align}
\begin{split}
    \Lambda_{\otimes} &= (J_{\otimes} A^{-1} J_{\otimes}^{T})^{-1} \\
    \mu_{\otimes} &= \overline{J}_{\otimes}^{T} b - \Lambda_{\otimes} \dot{J}_{\otimes} \dot{q} \\
    p_{\otimes} &= \overline{J}_{\otimes}^{T} g \\
    F_{\otimes} &= \overline{J}_{\otimes}^{T} \Gamma \\
    \overline{J}_{\otimes}^{T} &= \Lambda_{\otimes} J_{\otimes} A^{-1} 
\end{split}
\end{align}

The commanded whole-body control torque is $\Gamma = J_{\otimes}^{T} F_{\otimes}$, where $F_{\otimes}$ is the control force vector to control the tasks.

\subsection{Postural Synergies for Motion Generation in SynSculptor}

We propose \acro as a framework for humanoid motion generation based on humanoid postural synergies, compact representations that parameterize full-body motion as differential joint motions encoded in low-dimensional spaces. We compress temporal motion data by decomposing each pose and velocity trajectory
\(\{\mathbf{q}(t),\,\dot{\mathbf{q}}(t)\}_{t=0}^{T}\)
into synergies, where each synergy is defined by an initial configuration
\(\mathbf{q}_0\)
and a small set of principal joint–velocity vector directions
\(\{\dot{\mathbf{q}}_i\}_{i=1}^k\)
that capture the dominant temporal gradients in \(\dot{\mathbf{q}}(t)\).

To extract meaningful synergies from continuous motion, we segment trajectories based on momentum discontinuities. The system momentum is computed as
\begin{equation}
\mathbf{p}(t) = A(\mathbf{q}(t)) \dot{\mathbf{q}}(t),
\end{equation}
A new motion primitive is initialized whenever a significant momentum change is detected:

\begin{equation}
\| \mathbf{p}(t_i) - \mathbf{p}(t_{i-1}) \| >  \Delta P_{\rm th}
\end{equation}

\begin{figure}[h]
    \centering
    \includegraphics[width=0.98\linewidth]{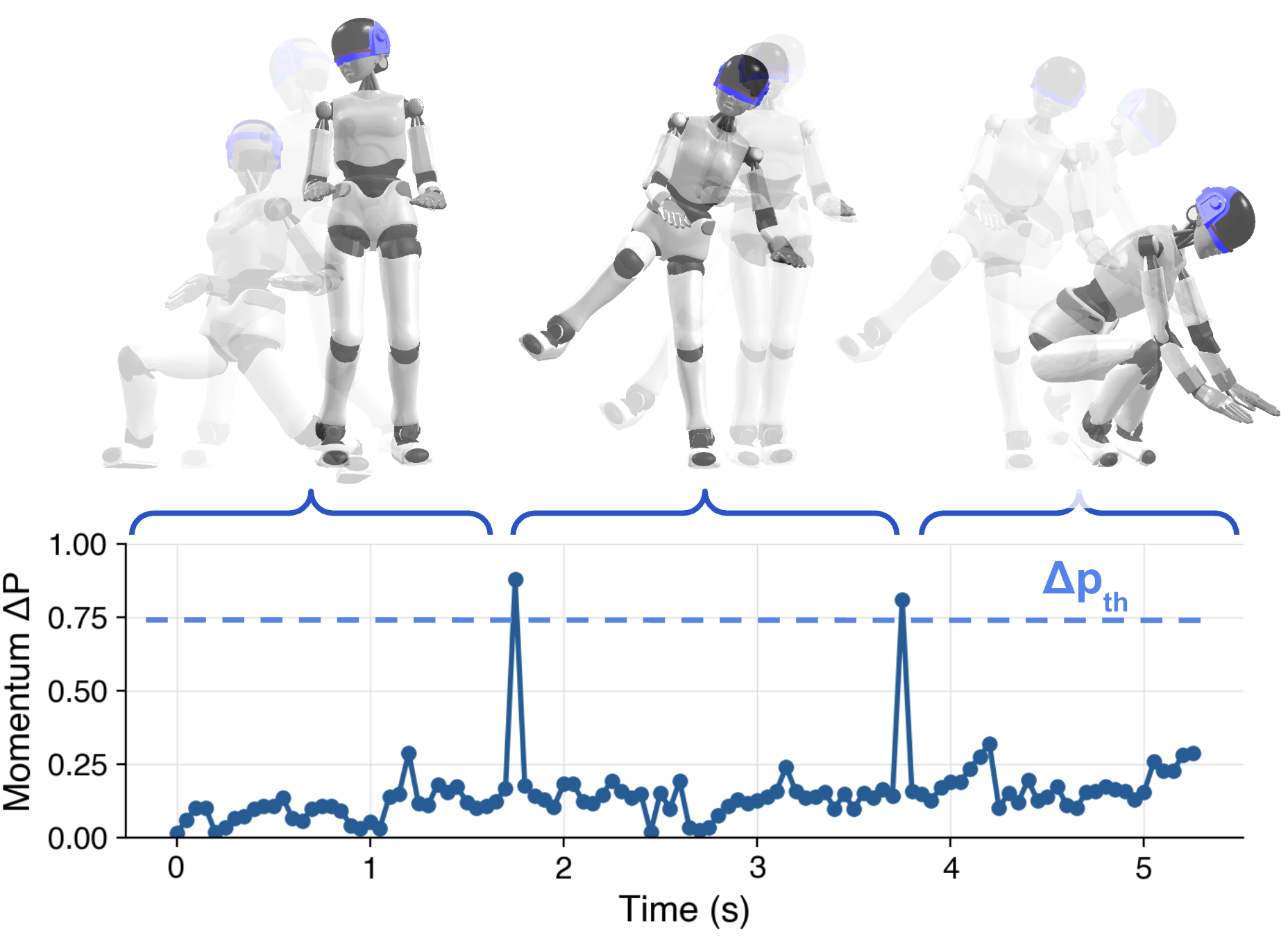}
    \caption
    {\textbf{Momentum‐Based Segmentation:} We compute the instantaneous change in whole‐body momentum \(\Delta P(t)\) at 100 Hz. A horizontal dashed line indicates the detection threshold \(\Delta P_{\rm th}\); when \(\Delta P(t)>\Delta P_{\rm th}\) (e.g., near 2 s and 4 s), the algorithm identifies major dynamic events such as foot‐lift or posture transitions. Between these peaks, \(\Delta P(t)\) remains low and noisy, reflecting only minor postural adjustments.}
    \label{fig:momentum-method}
\end{figure}

We set this threshold \(\Delta P_{\rm th} = 0.75 \), where reducing \(\Delta P_{\rm th}\) yields finer segmentation at the expense of diversity of motion explained by a synergy. This criterion produces segments of roughly 1–2 s for an average squatting motion, for example (Figure \ref{fig:momentum-method}).

Within each segment, a synergy is defined by a reference pose \( \mathbf{q}_0 \) and a low-dimensional velocity basis obtained via PCA on the joint velocity trajectories. Let $\widehat{\dot{\mathbf{q}}}(t)$ denote the reconstructed joint velocity; we approximate the true velocity $\dot{\mathbf{q}}(t)$ by a linear combination of the principal three synergy components, given upwards of 90\% reconstruction in all genres:
\begin{equation}
\widehat{\dot{\mathbf{q}}}(t)
\;=\;
\sum_{i=1}^{3} a_i(t)\,\dot{\mathbf{q}}_i
\;\approx\;
\dot{\mathbf{q}}(t),
\label{eq:vel_recon}
\end{equation}
where $\{\dot{\mathbf{q}}_i\}_{i=1}^3$ are the highest-variance PCA modes and $a_i(t)$ are scalar synergy coefficients, default to constant corresponding singular values for optimal (at least 90\%) reconstruction. \acro exposes those singular‐value coefficients as adjustable parameters for real‐time motion tuning. A full-body configuration is then recovered by integrating the reconstructed velocity:
\begin{equation}
\widehat{\mathbf{q}}(t)
\;=\;
\mathbf{q}_0
\;+\;
\int_{0}^{t} \widehat{\dot{\mathbf{q}}}(\tau)\,d\tau.
\label{eq:pose_recon}
\end{equation}

\acro enables real-time synthesis and editing of complex whole-body movements through time-varying coefficients. Users can import, reorder, and sequence entire synergy libraries via diverse interpolation schemes to customize style on the humanoid. This is critical for on-the-fly stylistic humanoid tuning and design, without requiring retraining or low-level retargeting. Our approach generates a broad range of coordinated, expressive motions while preserving style for modular and editable motion. The process to generate different motions through coefficient tuning is intuitive, as each coefficient is associated with a distinct synergy. The editor does not require large memory storage, as the synergies are compact.

\section{EXPERIMENTS}
We conduct a variety of experiments to validate the intuition and utility of human synergies and \acro by addressing the following objectives:

\begin{itemize}
\item Benchmark biomechanical vs robot baseline mechanical power consumption, validating motion mapping efficiency and motivating synergies broadly.
\item Identify the minimal synergy basis necessary to accurately reconstruct prototypical tasks and quantify stylistic expressivity across dance genres.
\item Evaluate reconstruction fidelity of the synergy basis by comparing the energetic profiles of reconstructed motions against original movements.
\item Investigate integration of human synergies into text-to-motion models to fine-tune stylistic outputs.
\end{itemize}

\subsection{Human vs.\ Robot Power Consumption}

To validate that our OptiTrack-based motion mapping and OpenSai control pipeline produces physically realistic humanoid behaviors, we compare the mechanical power consumption of mapped robot motions against human biomechanics. Human motions are simulated in OpenSim using the Rajagopal full-body model \cite{rajagopal2016}, which features 37 DOF, 80 Hill-type muscle–tendon actuators in the legs, and 17 ideal torque actuators for the torso and arms.  Humanoid counterparts are synthesized via our synergy-based control plus inverse kinematics.  For each, we compute instantaneous mechanical power and integrate over standard movement cycles (normalized by body mass).



For a multi-muscle musculoskeletal system in OpenSim, forward kinematics analysis computes each actuator’s instantaneous power \(P_j\) (in Watts), where positive values indicate energy delivery to the model and negative values indicate energy absorption. Summing over all \(n\) actuators yields the total human mechanical power:
\[
P_{\mathrm{human}} = \sum_{j=1}^{n} P_{j}.
\]

From our motion‐mapping framework, we compute the instantaneous mechanical power for the humanoid (with $m$ actuated joints) as
\begin{equation}
P_{\mathrm{robot}}
\;=\;
\sum_{i=1}^{m} \, \bigl|\tau_{i}\,\dot q_{i}\bigr|
\label{eq:robot_power}
\end{equation}

\begin{figure}
    \centering
    \includegraphics[width=0.95\linewidth]{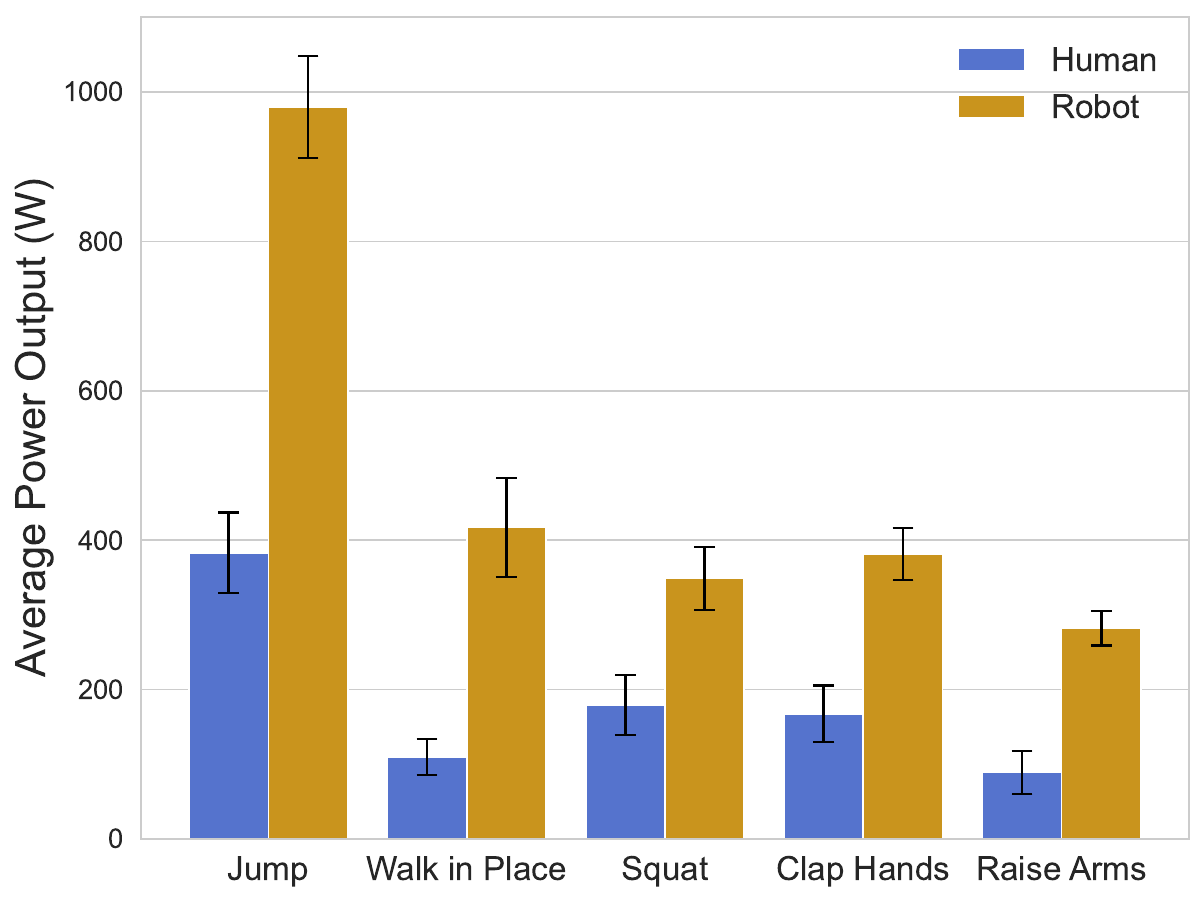}
    \includegraphics[width=0.95\linewidth]{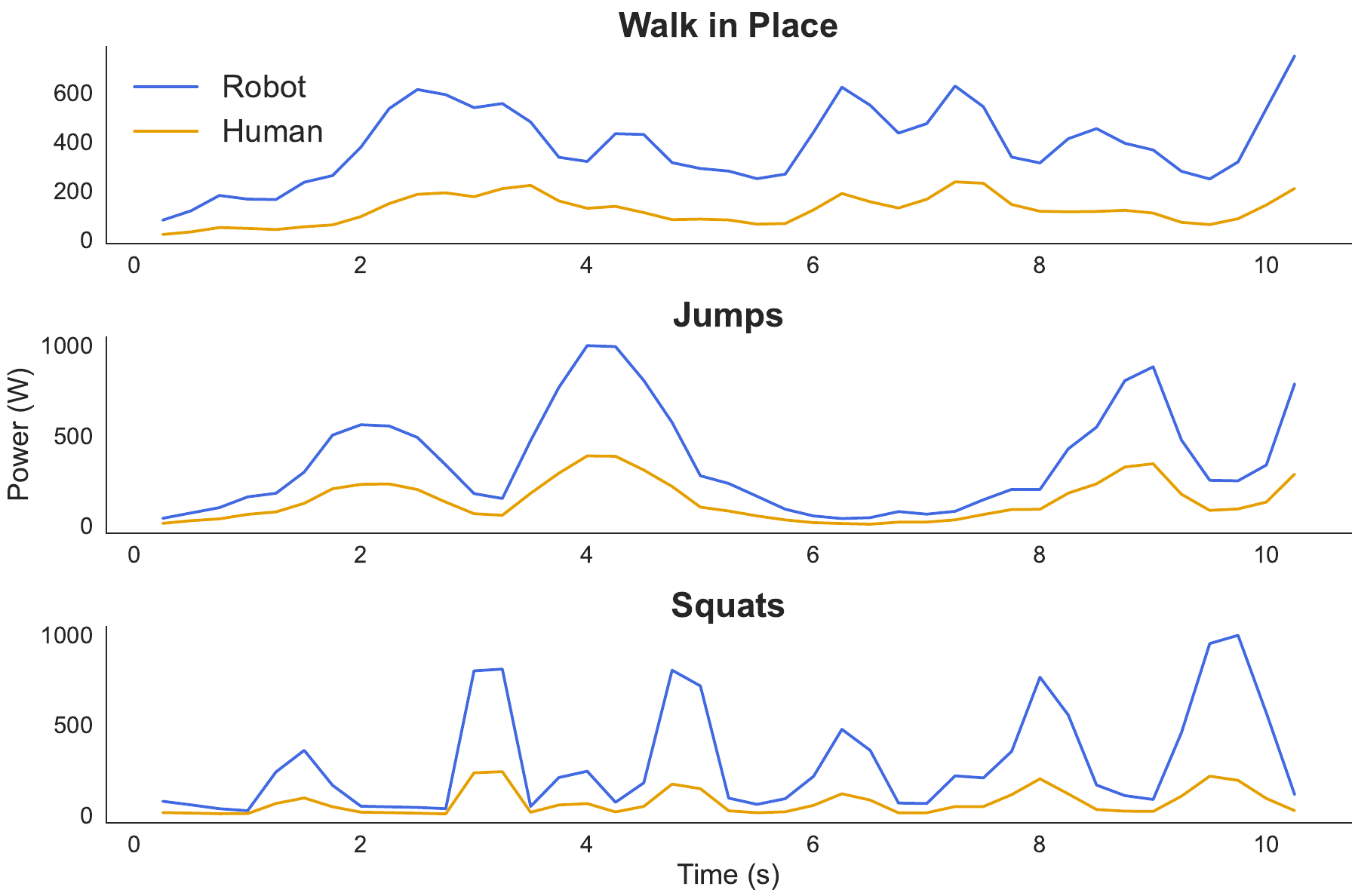}
    \caption{\textbf{Human and Robot Power Consumption}. We map MoCap data to a biomechanical model in OpenSim and our humanoid mapping framework to compute and compare power. (Top) Mean power consumption over 10-second trials, averaged across 20 individuals for each motion. (Bottom) Representative power time series for a single individual performing each motion.}
    \label{fig:human-robot-comp}
\end{figure}

We average \(P_{\mathrm{robot}}\) over a complete motion cycle and normalize by mass to obtain an energy‐per‐tracking cycle metric directly comparable to the human measure. Across 20 subjects and four canonical tasks, we find that human muscles operate, on average, 3.3× more efficiently than the corresponding robot actuators under direct motion mapping.  High-intensity behaviors (e.g.\ jumping, walking in place, squatting) show the largest human–robot efficiency gaps, driven by human synergies that exploit intrinsic muscle dynamics and tendon elasticity for balance and load transfer. In contrast, purely upper-limb tasks (e.g.\ clapping, arm raises) incur minimal power, with both platforms performing comparably (Figure~\ref{fig:human-robot-comp}).  
These results confirm that our mapping and control framework reproduces an anticipated efficiency penalty in robot mapping, while preserving physical realism. Biomechanical power experiments show that humans inherently exploit optimized musculoskeletal coordination beyond direct joint-by-joint mapping, further motivating the adoption of synergies. The OpenSim simulation for the human model and OpenSai simulation for the humanoid robot are both dynamics-based simulations using floating-base models while disregarding contact forces, which allows for a direct comparison between the two.

\subsection{Postural Synergies of Prototypical Movements}

We characterize the intrinsic dimensionality of human free-space dance motion using principal components (PCs) of momentum-parsed joint-velocity trajectories. Here, we highlight typical PC modes and their contributions to prototypical motions.

\begin{figure}[h]
    \centering
    \includegraphics[width=0.98\linewidth]{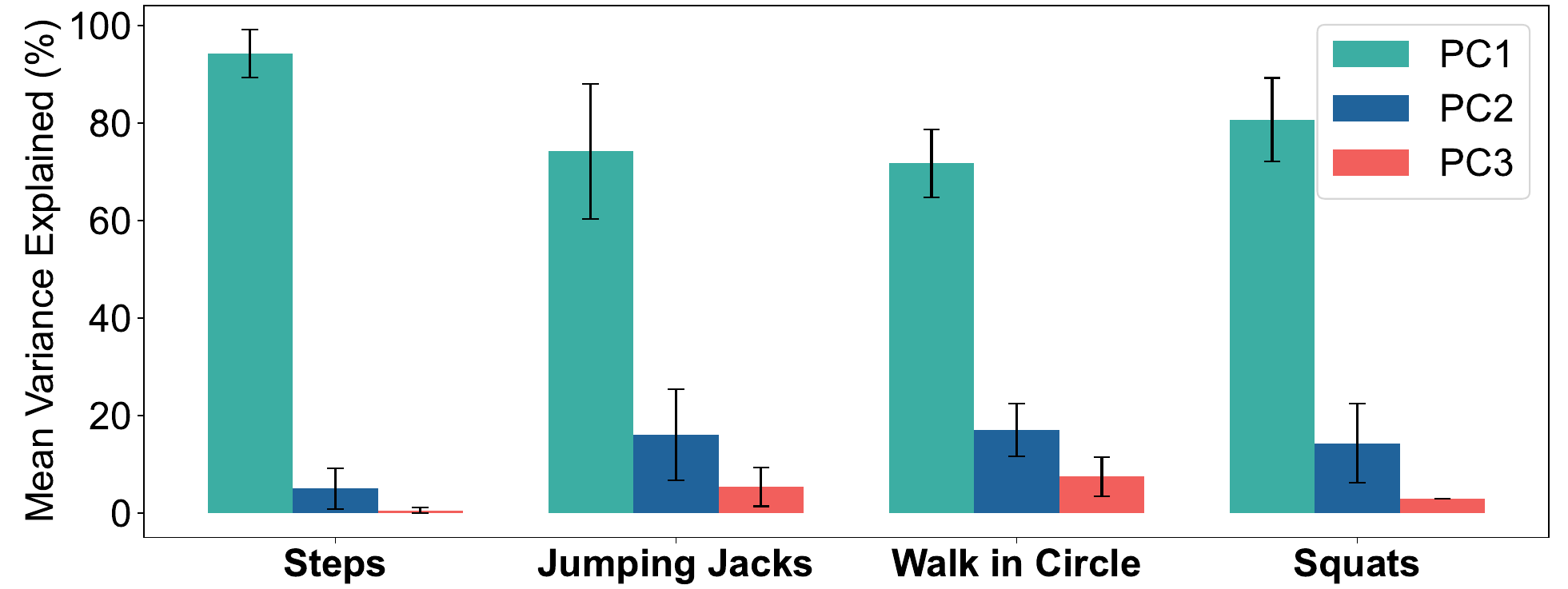}
    \caption{\textbf{Synergy Representation of Prototypical Movements:} We quantify inter‐subject variability for another set of four representative moves, over a longer time range (each performed for 30s by 20 participants). Steps in Place shows minimal dispersion, while Walk in Circle and Jumping Jacks require a tertiary synergy that contributes roughly 10\% of residual variance.}
    \label{fig:4}
\end{figure}

We recorded 30-second motion-capture trials of squatting, walking in a circle, stepping in place, and jumping jacks across 20 participants. The first principal component (PC1) dominates in highly repetitive, single-axis movements (stepping in place and squatting), whereas the relative contributions of PC2 and PC3 increase for movements requiring orthogonal or asymmetric limb coordination (jumping jacks and circular walking). Error bars on each component’s mean variance explained quantify the consistency of synergy recruitment over successive movement cycles, with larger spreads indicating greater shifts in coordination strategy over time (Figure \ref{fig:4}).

\paragraph{Stepping in Place}
Stepping in place exhibits 94–100\% variance explained by the first principal component, demonstrating that this highly stereotyped cyclic movement is effectively captured by a single synergy per segment. The negligible contributions of PC2 and PC3 indicate minimal deviation from the dominant pattern, with each step executed in nearly identical mechanical and spatial form across cycles.

\paragraph{Jumping Jacks}
Jumping jacks require concurrent oppositional motions of the upper and lower limbs, introducing at least two orthogonal coordination modes. The elevated contributions of PC2 and PC3 correspond to these distinct degrees of freedom, for example lateral versus vertical limb displacement. Greater variance across participants and segments indicate the relative weighting of these synergies fluctuates during the transitions between "open" and "closed" configurations.

\paragraph{Walk in Circle}
Walking in a circular path adds directional changes and asymmetric foot placements. Although PC1 continues to capture the core gait rhythm, PC2 and PC3 register the turning-induced lateral shifts and trunk realignments. The moderate error bars on these secondary components qualitatively reflect variability between tighter and wider turns across segments.

\paragraph{Squatting}
Squats consist predominantly of vertical hip and knee flexion–extension, so a single synergy accounts for the bulk of the variance. PC2 encodes smaller control nuances, such as subtle trunk lean or ankle dorsiflexion, whereas PC3 remains negligible.

\subsection{Stylistic Variability across Dance Synergies}
\begin{figure}[h]
    \centering
    \includegraphics[width=0.98\linewidth]{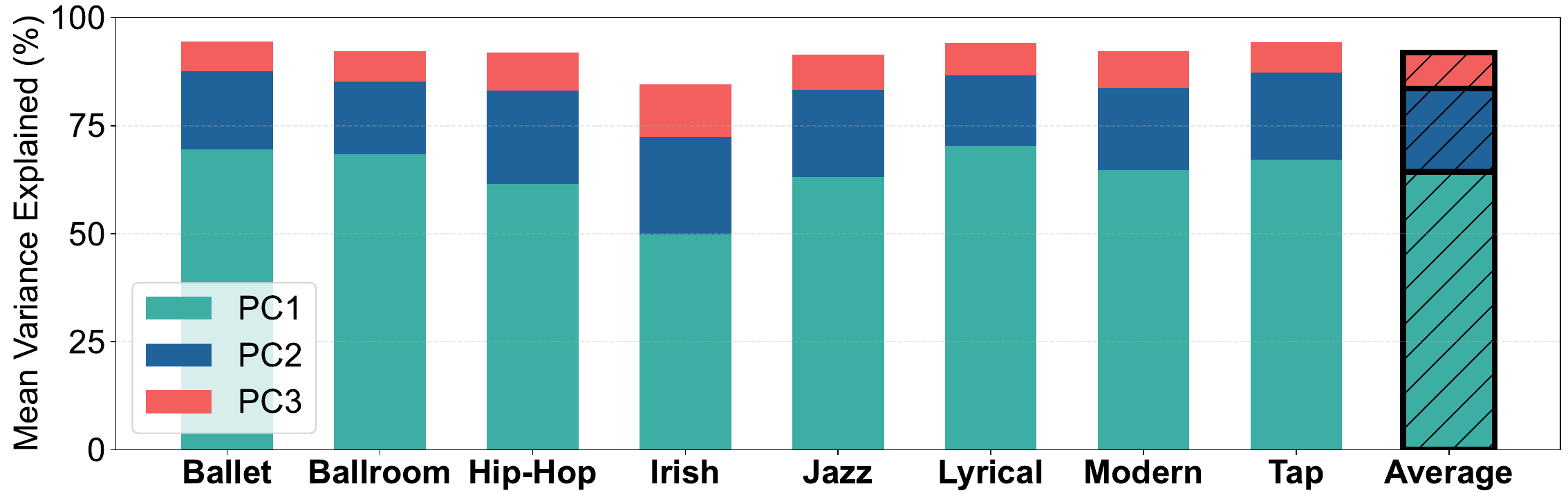}
    \caption{\textbf{Stylistic Generalization Across Dance Genres.} The span of momentum-segmented synergies collectively capture over 90\% of total variance in all examined genres. Moreover, the second and third components contribute on average 30\% and 13\% of the variance explained by the primary synergy.}
    \label{fig:5}
\end{figure}

We evaluate genre-specific expressivity across eight dance styles by projecting each style’s motion‐capture data into a shared low‐dimensional synergy subspace and measuring the variance retained by this compression. Across all eight dance genres, within data from a single dancer, the first three principal components capture on average 64.3\%, 19.3\%, and 8.3\% of total variance, respectively (Figure \ref{fig:5}). This confirms that a subspace with 3 basis vectors suffices, even for radically different styles.

Ballet and Lyrical synergies confine most variance to their primary synergy (72–74 \%), with the second and third components together accounting for only 25\% of variance. Classical-based genres thus yield highly stereotyped coordination patterns, whereas more contemporary and folk forms distribute variance more evenly across multiple synergies. These inter‐genre distinctions suggest that synergy‐based controllers may require genre‐specific weighting to reproduce the stylistic richness of disciplines like Irish dance and Hip-Hop, while simpler low‐dimensional mappings would suffice for Ballet and Lyrical dance motions.

Hence, a fixed 3-D synergy basis not only compresses all eight genres above 90\% fidelity, it also highlights which styles need richer secondary and tertiary modes. In the prototypical movements dataset, three components collectively explain on average $96\%$ of the joint‐velocity variance; thus, scaling such linear combinations in $\dot{q}$-space suffices to synthesize diverse and high‐fidelity motion variations.

\subsection{Motion Generation}
\begin{figure}[htbp]
    \centering
    \includegraphics[width=0.98\linewidth]{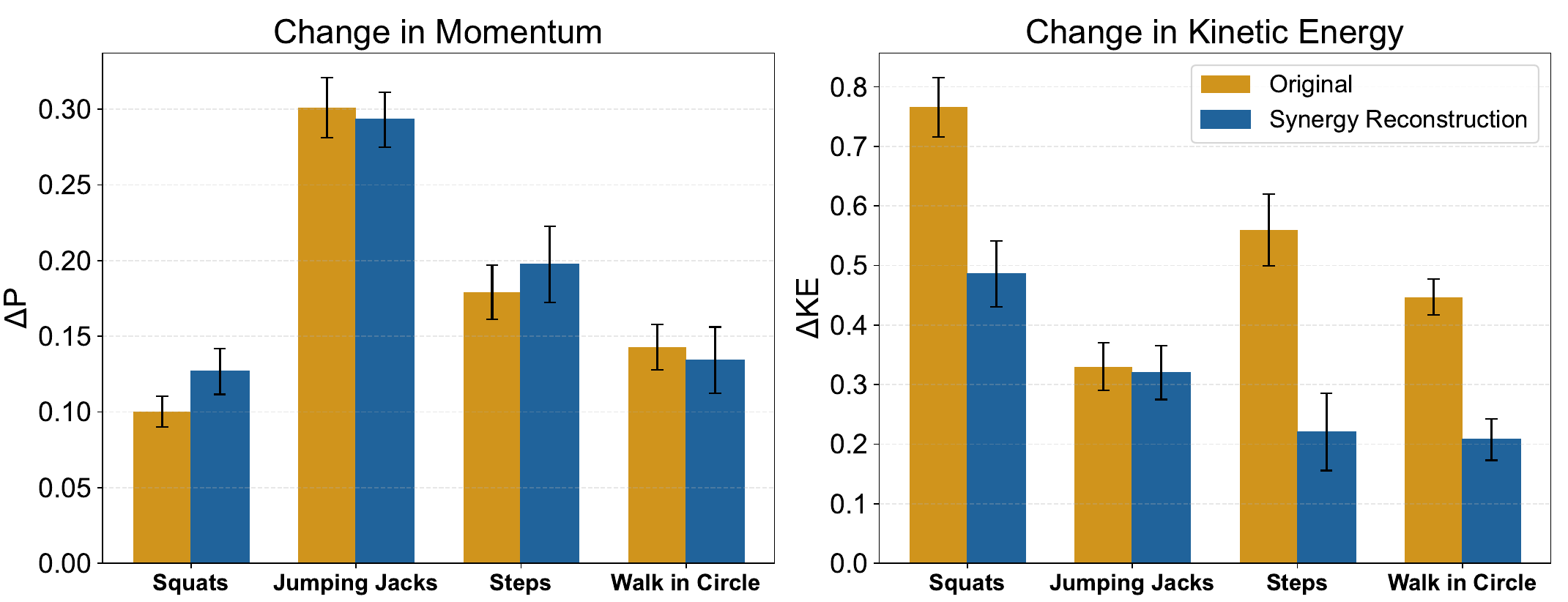}
    \caption{\textbf{Energetic Reconstruction.} Comparison of mean instantaneous changes in whole‐body momentum ($\overline{\Delta P}$) and kinetic energy ($\overline{\Delta KE}$) at 1 kHz for four prototypical moves.  Synergy‐based reconstructions (100 random samples in the 3D synergy subspace) closely match the original energetic profiles, demonstrating that the reduced basis preserves core dynamics while filtering out high‐frequency components.}
    \label{fig:energetic_recon}
\end{figure}

We assess motion similarity by comparing the synthesized and original trajectories’ mean $\Delta P$ and mean $\Delta KE$. These quantify average frame-to-frame momentum and kinetic energy fluctuation, as proxies for motion speed smoothness and dynamic consistency or jitteriness, respectively.

To verify that our synergy basis can reproduce the energetic characteristics of the original motions, we perform a Monte Carlo synthesis over each segmented “move.”  For each prototype, we draw 100 random coefficient vectors 
\(\mathbf{a}\sim\mathcal{U}(-\alpha,\alpha)^3\) in the subspace spanned by the top three PCA directions \(\{\dot{\mathbf q}_i\}\).  The reconstructed joint‐velocity is given in Eq.~(\ref{eq:vel_recon}), which we integrate to recover the full configuration in Eq.~(\ref{eq:pose_recon}).
We then compute the mean per‐step values \(\overline{\Delta P}\) and \(\overline{\Delta KE}\) at a control rate of 1 kHz.  For each trajectory, instantaneous energetic metrics are defined as
\begin{align}
    \Delta P(t_{k}) = \bigl\|\mathbf p(t_{k})-\mathbf p(t_{k-1})\bigr\|
\end{align}
where $\mathbf{p}$ is the momentum and 
\begin{align}
    \Delta KE(t_{k}) = \bigl|KE(t_{k}) - KE(t_{k-1})\bigr|
\end{align}
where $KE = \frac{1}{2} \dot{\mathbf{q}}^{T} A \dot{\mathbf{q}}$ is the kinetic energy. These metrics encoder motion similarities in terms of momentum discontinuities and energy injection and removal.

As shown in Figure \ref{fig:energetic_recon}, momentum trajectories generated from randomly weighted synergy reconstructions virtually overlay the original profiles, confirming that our three‐dimensional synergy subspace preserves each movement’s core dynamics. The mean kinetic‐energy variation $\overline{\Delta KE}$ in synergy reconstructions is reduced by roughly 32\%, indicating smoother trajectories by excluding minor synergies.

\subsection{Synergy-Augmented MotionGPT with Posture Control}

We extend our synergy framework to enhance MotionGPT’s text-driven motion generation for humanoid control. MotionGPT \cite{jiang2023} autoregressively predicts 3D human pose sequences via a learned “motion vocabulary,” enabling text-to-motion and motion-to-text. However, direct decoding of MotionGPT tokens onto a floating-base humanoid produces physically implausible behaviors—foot slip, joint jitter, and loss of balance—due to its high-dimensional, unconstrained output space.

To enforce human-like, natural movement, we project MotionGPT’s raw joint‐velocity output into the synergy subspace while removing any torso motions. By using the nullspace matrix associated with the torso task:
\begin{align}
N_{t} = I - \overline{J}_{t} J_{t}
\end{align}
then for MotionGPT’s raw output $\dot{\mathbf q}_{\mathrm{GPT}}(t)$, we compute
\begin{align}
\widehat{\dot{\mathbf q}}_{\, GPT | t}(t)
\;=\;
S\,S^{T}\,N_{\mathrm{t}}\,\dot{\mathbf q}_{\mathrm{GPT}}(t)\,
\end{align}
where $S$ is the synergy basis, ensuring that the resulting velocity lies both in the learned synergy subspace and in the nullspace of the torso motion. We overwrite the operational‐space derived torso torque commands with that reconstructed from the synergy‐projected velocities, ensuring that torque control remains consistent with the learned synergy representation. 

The generated torso motion from SynSculptor is added directly to the projected MotionGPT velocities, resulting in the following whole-body joint velocity trajectory:
\begin{align}
    \widehat{\dot{\mathbf q}}(t)
\;=\;
\widehat{\dot{\mathbf q}}_{\, GPT | t}(t) + \widehat{\dot{{\mathbf q}}}_{\,t}(t)
\end{align}

To analyze realism and kinematic efficiency, we report both mechanical power and a foot-sliding ratio, in which a foot either penetrates or slides along the virtual ground plane, flagging a violation of realistic contact behavior in our floating‐base simulation, adopted from Cohan, et. al \cite{cohan2024}.
 
\begin{figure}
    \centering
    \includegraphics[width=0.98\linewidth]{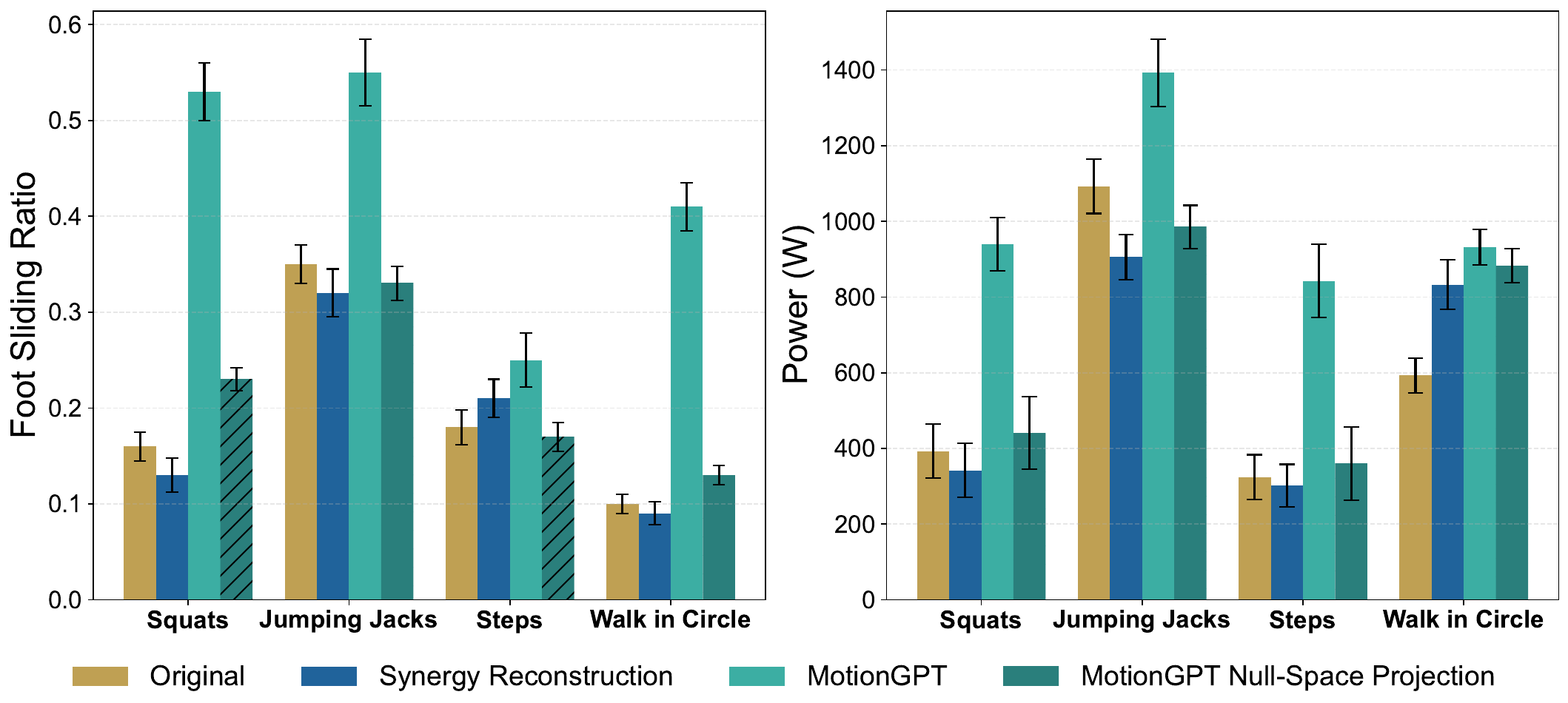}
    \caption{\textbf{MotionGPT Fine-Tuning via Null-Space Synergies.}  
    20 MotionGPT trials per movement reveal that raw MotionGPT outputs (teal) exhibit elevated foot-sliding ratios and power compared to original MoCap mappings (gold) and synergy-only reconstructions (blue). Projecting MotionGPT trajectories through the torso null-space synergy reduces both foot slipping and power consumption, yielding performance on par with original mappings.}
    \label{fig:motiongpt}
\end{figure}

Projecting MotionGPT’s outputs through our torso null‐space synergy reduces the foot‐sliding ratio by an average of 20–35\% across four prototypical tasks, substantially improving contact realism. Furthermore, this synergy‐based projection lowers instantaneous mechanical power demand—computed via the joint‐torque velocity inner product—by up to 54\% during squats, evidencing markedly enhanced kinematic efficiency and closer adherence to human movement patterns (Figure \ref{fig:motiongpt}).

Synergies provide a powerful low-dimensional inductive bias for diffusion-based generative policies, allowing text-conditioned motion outputs to be projected directly onto humanoid null-spaces. This integration bridges the gap between large-scale text-to-motion models and physically and stylistically accurate robot execution.

\section{CONCLUSION}
Synergies serve as shareable building blocks that a humanoid can recombine to perform movements never explicitly trained on or demonstrated. We coordinate joints through low-dimensional synergies for smoother, more coordinated trajectories, biasing motion toward human-like efficiency. Empirically, synergy-driven motions in \acro exhibit closer kinematic alignment with human data, such that embedding motion generation within a synergy space enables real-time stylistic modulation. In SynSculptor, genre-specific dances, expressive gestures, and mood-driven motions are realized as variations along synergy axes, decoupling style from core kinematics. This real-time stylization opens the door to interactive humanoid performances and on-the-fly avatar personalization that have previously been challenging with offline motion blending approaches. Once extracted, these synergies can be reused to generate new motions without task-specific retraining of each new motion skill. By constraining generative models to operate in synergy space, we inject a strong prior for biomechanically realistic coordination. 

\section{LIMITATIONS AND FUTURE WORK}
While \acro has demonstrated the potential of low-dimensional synergy spaces for expressive humanoid control, our current pipeline focuses on kinematic reconstruction of human motions, but does not guarantee dynamic stability or contact feasibility. To deploy \acro\ on a bipedal platform, the synergy planner must be augmented with humanoid balance controllers that actively modulate center‐of‐mass dynamics to prevent falls.
We have demonstrated that synergies can compactly encode complex motion styles, such that future work should focus on synergy-conditioned synthesis and generating novel, style-consistent trajectories in the low-dimensional synergy space. Moreover, \acro can be extended to support conditioned interpolation between synergies through diffusion-based pose blending.

\section*{ACKNOWLEDGMENT}

We gratefully acknowledge dancers Maya Alvarez-Coyne, Hannah Woolfden, and Tracy Paige. This work was supported by the Stanford Robotics Center.

\end{document}